# VAE-iForest: Auto-encoding Reconstruction and Isolation-based Anomalies Detecting Fallen Objects on Road Surface

Takato Yasuno[*1]  Junichiro Fujii[*1], Riku Ogata[*1], Masahiro Okano[*1]

[*1] Yachiyo Engineering, Co., Ltd., RIIPS.

In road monitoring, it is an important issue to detect changes in the road surface at an early stage to prevent damage to third parties. The target of the falling object may be a fallen tree due to the external force of a flood or an earthquake, and falling rocks from a slope. Generative deep learning is possible to flexibly detect anomalies of the falling objects on the road surface. We prototype a method that combines auto-encoding reconstruction and isolation-based anomaly detector in application for road surface monitoring. Actually, we apply our method to a set of test images that fallen objects is located on the raw inputs added with fallen stone and plywood, and that snow is covered on the winter road. Finally we mention the future works for practical purpose application.

## 1. Introduction

### 1.1 Related Works for Road Surface Recognition

Since 2019, the focus has always been on an intelligent system for monitoring road surface conditions using computer vision and machine learning [Carrillo 2019–Yasuno 2022]. For example, Grabowski et al. proposed an intelligent road sign system including a detector using convolutional neural networks (CNNs) and image processing video cameras for the classification of dry, wet, and snowy roads [Grabowski 2020]. Liang et al. investigated the application of semantic segmentation network customized with dilated convolutions for road surface status recognition [Liang 2019]. Though they have succeeded to improve the performance of classification and segmentation architectures for road surface recognition; however, the daily important issue has not be solved where some hazardous objects might be fallen on the road surface for road managers and users.

Since 2018, the several articles have been found on the digital sensing opportunities and fallen rocks detection challenges on transportation road surface. Amini et al. proposed an application using the variational auto-encoder for control autonomous driving with training de-biasing [Amini 2018]. Jung et al. proposed an application of noise-based road surface anomaly detection for efficient non-compression auto-encoder for driving [Jung 2021]. Abdelmaboud et al. proposed a method of rock-fall risks reduction for early warning system [Abdelmaboud 2021]. Shi el al. tried to detect small-scale fallen rocks on transportation roads using lidar point clouds [Shi 2021]. However, these methods could not completely solved to provide an end-to-end pipeline from noise reduction to hazardous score alert to detect fallen rocks on the road surface safely for users.

### 1.2 Fallen Object Detection on Road Surface

As illustrated in Figure 1, we propose an application to provide anomaly score information with fallen objects hazard alerts using live camera images. The increasing usage of CCTV and IP cameras has opened the possibility of using them to automatically detect hazardous road surfaces and inform drivers through alerts. In daily road monitoring, it is an important issue to detect changes in the road surface at an early stage to prevent damage to third parties. The target of the fallen object may be a fallen tree due to the external force of a flood or an earthquake, and fallen rocks from a slope. Especially in winter we aim to provide a new snow hazard index on the amount of snow covered on the road surface. During a heavy snowfall, the road surface region is covered with snow and is not clearly visible on the live camera. Through live image, the background snow excluded with road region would not influence road user for snow hazard. Thus, we need to detect whether snow has covered on the road surface.

We prototype a method that combines auto-encoding algorithm and anomaly detection to detect fallen objects on the road surface using the road monitoring images. Actually, we apply our method to a set of test images in which some fallen objects of the road surface is located on several experimental images such as fallen stone, plywood board, and snow. Furthermore, we mention the future works for practical purpose application and the usefulness.

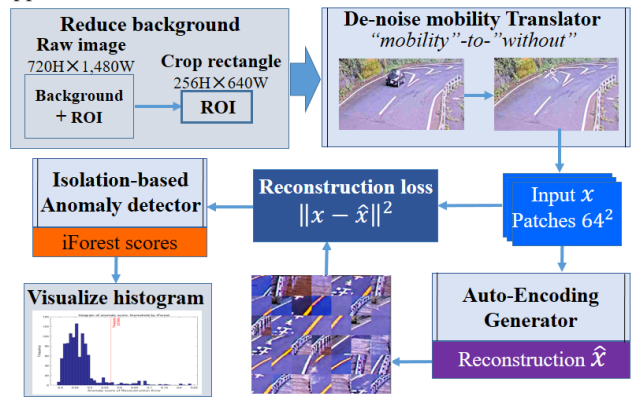

Figure 1: Overview of our pipeline to detect fallen objects

## 2. Building Fallen Object Detector

### 2.1 Reduce Background and Mobility Region

The image-to-image translation is possible for training a paired image dataset under the fixed angle of road camera. Let us note that the background region without road surface could become unstable representation learning. Because tree is variable owing to seasonal color, strong wind. We propose the pre-process to

Contact: Takato Yasuno, RIIPS on 3F, 5-20-8, Asakusabashi, Taito-ku, Tokyo, 111-8648, tk-yasuno@yachiyo-eng.co.jp

"crop rectangle" covering on the region of road surface. In addition, some mobility, i.e. bus, truck, motorcycle, bicycle, sometimes are moving on the road surface. We propose a method to generate the road surface to reduce the "*with-mobility*" region using L1-Conditional GAN. In the original pix2pix study, the input edge images were translated to shoe images [Isola 2017]. Ammar proposed a method to detect moving objects using segmentation and classification in video surveillance [Ammar 2015]. However, their study was not any de-noise method to delete the region of moving mobility. We apply a generative deep learning method, in which the *with-mobility* on the road surface is translated into *without-mobility* fake output using the conditional GAN, pix2pix.

### 2.2  Auto-Encoding Application for Road Surface

The past decade has seen a renewed importance in the representation learning applications for anomaly detection, born in medical industry [Schlegl 2019]. Sabuhi et al. presented a systematic literature review of the applications of generative adversarial networks (GANs) in anomaly detection, covering 128 papers [Sabuhi 2021]. In contrast, Kingma et al. proposed the variational auto-encoder (VAE) embedding the bridge of latent space using re-parameterization trick [Kingma 2013]. In addition, the adversarial auto-encoder (AAE) has proposed in the application on X-ray security imagery [Akcay 2018]. Furthermore, the VAE-GAN combined approach proposed and experimental studied in MNIST, Coil-100 dataset [Bian 2019].

These auto-encoding algorithms directly enable to compare the raw input with anomalous feature and the reconstructed fake output for anomaly detection. However, it is not widely understood whether the auto-encoding algorithm can practically contribute to detect anomalous feature. We focus on the auto-encoding algorithms such as the AAE, VAE, and cVAE [Zhang 2021]. This study applies auto-encoding methods to detect fallen objects on the road surface monitoring. These methods lead to the practical purpose application for live camera monitoring of road surface. We demonstrate these auto-encoding methods for detecting plywood, fallen stone, and snow.

### 2.3  Isolation Metrics by Reconstruction Loss

There are several proposal and experimental studies on the anomaly detection using the auto-encoders [Norlander 2019–Ulger 2021]. There are many experimental metrics to detect anomalies whose subgroups are categorized into *reconstruction loss, lower bound gradient loss,* and *excess mass*. For examples, excess mass, mass volume [Norlander 2019], the KL-divergence, the average norm of reconstruction loss divided by the variance of decoder output [Pol 2020], gradient loss that stands for the derivative of the lower bound (ELBO), the combined metrics of reconstruction loss and weighted gradient loss [Ulger 2021]. However, it is not widely known which anomaly scores can accurately compute as the best metrics of anomalies.

The authors use the reconstruction loss between the raw road surface image and the auto-encoding output as anomaly metrics for road surface. Furthermore, we use the reconstruction loss as the input into the *isolation-based* anomaly detection algorithm [Liu 2012]. Because the region of fallen objects is always narrow compared with the whole road surface; such as plywood and fallen stones. At first the snow region is able to be covered on the whole surface, meanwhile the road would be rutted after cars passed. Therefore, the region of fallen objects is almost isolated. The *isolation forest (iForest)* algorithm is a different model-based procedure that explicitly isolates anomalies, instead of profiles normal points. It has a linear time complexity with a low memory requirement. The isolation characteristic of *iTrees* enables them to build partial models and exploit sub-sampling, so that an iTree isolates normal points [Liu 2012]. The iForest algorithm has two step procedure: the training step constructs the isolation trees (iTrees) sub-sampling from the training set; the evaluation step calculates the anomaly score for each test set.

The authors use the reconstruction loss between the raw road surface input and the auto-encoding output as anomaly metrics. Furthermore, we use the reconstruction loss as the input into the *iForest* algorithm to compute the road surface anomaly socores.

## 3. Applied Results

### 3.1  Reduce Background and Mobility Region

Herein, we demonstrate that our pipeline could automatically compute a snow hazard ratio index at the cold region. Figure 2 illustrates the ground-truth images of the road surface with-mobility. Figure 3 depicts a road surface without-mobility fake output translated from a raw input using a trained pix2pix.

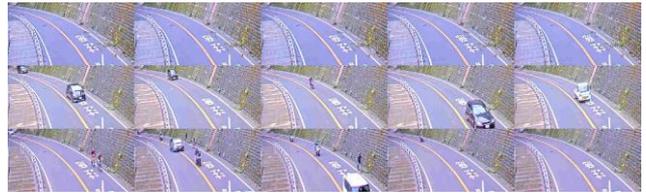

Figure 2: Raw inputs "*with-mobility*" added fallen stones, plywood.

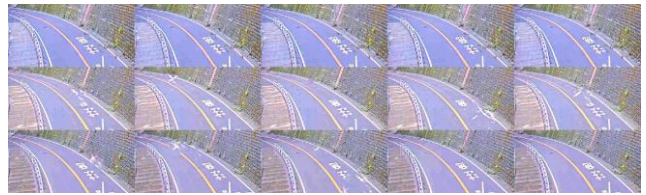

Figure 3: Translated outputs "*without-mobility*" remained fall objects.

We prepared 1,063 one-to-one datasets that included two groups: the *with-mobility* image and the *without-mobility* image. We collected live road images from Sept to Oct 2021 at mountainous road. The ratio of training data versus test data was 90:10. We set the input size of pix2pix network as $256H \times 512W$. We iterated 50 epoch using the Adam optimizer; therefore, it took 7 hours.

### 3.2  Plywood, Stone Detection on Road Surface

We prepared 532 training images and 43 test images by collecting an IP camera with size of $720 \times 1,480$ collected by an IP camera on the autumn season at Sept 2021. Here, we simulated test images that added the plywood and fallen stones by various scales over raw road images. We cropped images of $251 \times 650$ size for covering the region of road surface to reduce the background noise such as tunnel, poles, and tree. Furthermore, we resized $256 \times 640$ so as to divide completely without any remains and also extracted 4 × 10 = 40 unit patches with size 64 × 64, and selected the only 23 unit patches covering on the region of road surface. Therefore, the total number of training patches was 23 × 532 = 12,236. We set the input size of auto-encoding algorithms as same as the unit patch size 64 × 64. We set the number of latent space 128. We trained 400 epoch iterations with a mini-batch of 128 using the Adam optimizer; therefore, it took 107 minutes. On the other hand, the number of test patches was 1,000 that contains 82 anomalies of fallen stones and plywood on the road surface.

Table 1: Accuracy comparison of auto-encoding reconstruction mask and ground-truth segment of plywood and fallen stones.

|  | SSIM | Dice Similarity |
|---|---|---|
| **VAE** | **0.8115** | **0.6879** |
| cVAE | 0.8082 | 0.6869 |
| AAE | 0.7174 | 0.5238 |

Table 2: Accuracy of VAE-iForest fallen stones and plywood detector.

| | | Prediction | |
|---|---|---|---|
| | | normal | anomaly |
| **Actual** | normal | 908 | 34 |
| | anomaly | 10 | 48 |
| Recall | | 82.8% | |
| Precision | | 58.5% | |

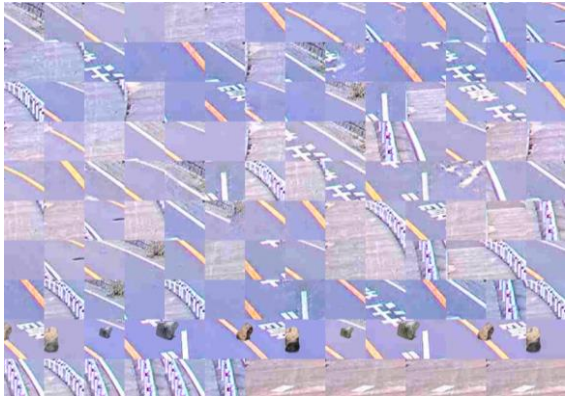

Figure 4: Raw input patches included the plywood and fallen stones.

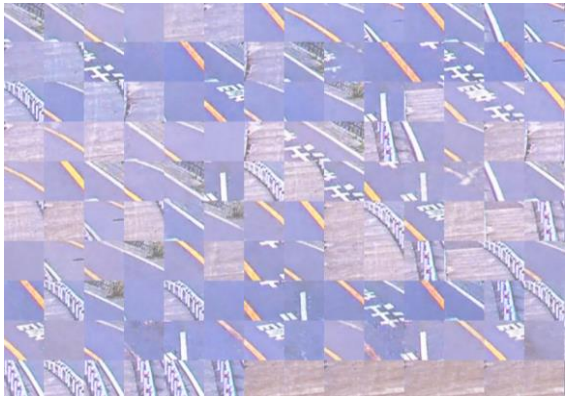

Figure 5: Reconstructed outputs predicted by VAE at Sept 2021.

Table 1 shows the accuracy comparison of the three auto-encoding reconstructed mask output that subtracted the ground-truth region of plywood and fallen stones. Among three auto-encoders, the Variational Auto-Encoder (VAE). has higher similarity than the Adversarial Auto-Encoder (AAE) and Conditional VAE (cVAE). Thus, the authors selected the VAE as the most practical auto-encoder in this experiments. As presented in Table 2, we depicted the accuracy of the VAE-iForest for fallen stones and plywood detector with recall and precision. As shown in Figure 4, we show the raw input patches included with the fallen objects such as plywood and fallen stones on the road surface. Figure 5 draw the reconstructed road surface outputs predicted by the VAE. As depicted in Figure 6, we draw a histogram of anomaly scores using the isolation forest algorithm to optimize a threshold to detect plywood and fallen stones on the road surface. Here, we set the fraction 0.04 that the total test images 1,000 contains 40 fallen objects on the road surface.

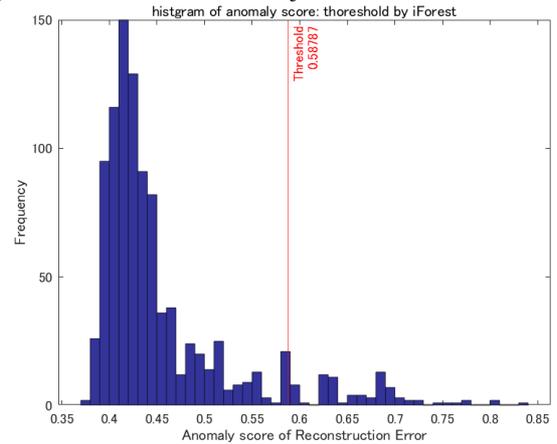

Figure 6: Histogram of anomaly scores using isolation forest algorithm to detect plywood and fallen stones on the road surface.

### 3.3 Snow Detection on Road Surface

We prepared 1,039 training images and 15 test images by collecting an IP camera with size of 720 × 1,480 collected by an IP camera on the winter season at Jan 2022. We prepared 2,070 raw test patches that contains snow region by various condition on the road images. We cropped images of 251 × 650 size for covering the region of road surface to reduce the background noise such as tunnel, poles, and tree. Furthermore, we resized 256 × 640 so as to divide completely without any remains and also extracted 4 × 10 = 40 unit patches with size 64 × 64, and selected the only 23 unit patches covering on the region of road surface. Therefore, the total number of training patches was 23 × 1,039 = 23,897. We set the input size of auto-encoding algorithms as same as the unit patch size 64 × 64. We set the number of latent space 128. We trained 100 epoch iterations with a mini-batch of 128 using the Adam optimizer; therefore, it took 60 minutes. On the other hand, the number of test patches was 2,070 that contains 345 anomalies of fallen stones and plywood on the road surface. As presented in Table 3, we depicted the accuracy of the VAE-iForest for snow detector with recall and precision. Figure 7 show the raw input patches snow covered road surface. Figure 8 draw the reconstructed road surface outputs predicted by the VAE. As depicted in Figure 9, we draw a histogram of anomaly scores using the iForest algorithm to optimize a threshold to detect snow region on the road surface. Here, we set the fraction 0.167 that the total test images 2,070 contains 345 snow region on the road surface.

Table 3: Accuracy of VAE-iForest snow detector on road surface.

|  |  | Prediction | |
|---|---|---|---|
|  |  | normal | anomaly |
| **Actual** | normal | 1,665 | 60 |
|  | anomaly | 60 | 285 |
| Recall | | 82.6% | |
| Precision | | 82.6% | |

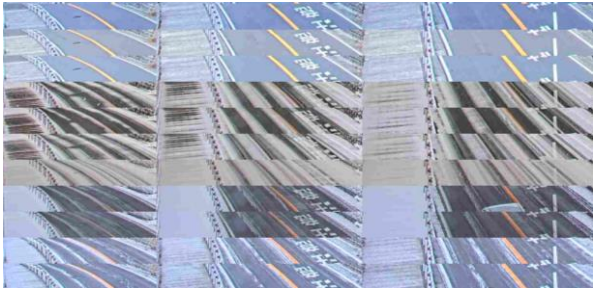

Figure 7: Raw input patches included the snow region.

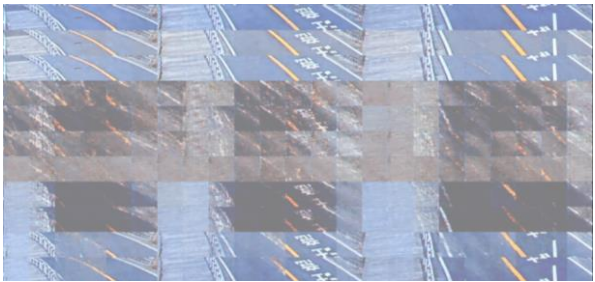

Figure 8: Reconstructed outputs predicted by VAE at Jan 2022.

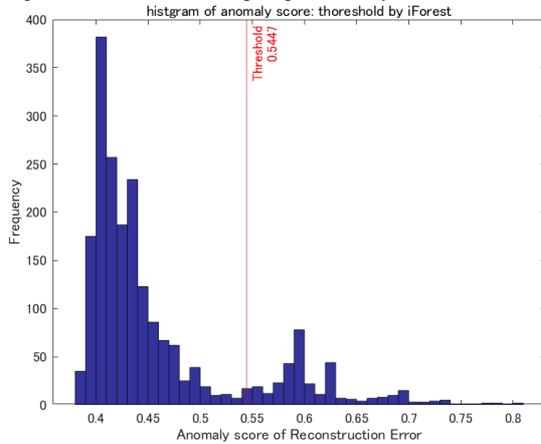

Figure 9: Histogram of anomaly scores using isolation forest algorithm to detect snow region on the road surface.

## 4. Concluding Remarks

This study built a prototype method that combines the variational auto-encoder (VAE) and anomaly detection (iForest) to detect fallen objects on the road surface using the road monitoring images. Actually, we apply our method to a set of test images in which several hazardous objects are located on the road surface. We annotated the simulated images added with the plywood and fallen stones by various size and collected the raw snow images for our experimental studies. As far as vision-based machine learning experiments, we found that our proposed method has 82 percent accuracy of recall for detecting fallen objects such as plywood, fallen stones, and snow region.

Furthermore, we mention the future works for more practical and robust application. We have experimental opportunities to reconstruct another road surface such as urban cross road for detecting fallen objects that maybe more frequently happen than the mountainous road and would be many types of hazardous fallen objects. In this case, not like the fixed point IP or CCTV camera, we have opportunities to use another dynamic point of drive record camera on daily patrol cars. Here, the front angle vision for the direction to drive a car are variable at each points on a rout. We could be able to learn flexibly using semantic segmentation algorithm to focus the region of road surface.

**Acknowledgments** The authors wish to thank the MathWorks, Takuji Fukumoto, Shinichi Kuramoto who provided us valuable help in the early studies of auto-encoding and anomaly detection.